\documentclass[11pt]{article}

\usepackage[preprint]{acl}

\usepackage{times}
\usepackage{latexsym}
\usepackage{fontawesome5}
\usepackage[export]{adjustbox}
\usepackage{tabularx} % Add in preamble
\usepackage{amssymb} % for \checkmark
\usepackage{hyperref}
\usepackage{url}
\usepackage{multirow}
\usepackage{bbm}
 \usepackage{makecell}
\usepackage{tikz}
\usetikzlibrary{positioning, arrows.meta, shadows, shapes.geometric}
\usepackage{pgfplots}
\pgfplotsset{compat=1.18}
\usepackage{graphicx}
\usepackage{amssymb}
\usepackage{amsmath}
\usepackage{cleveref}
\usepackage{amsthm}
\usepackage{booktabs}
\usepackage{lscape}
\usepackage{pifont}
\usepackage{multirow}
\usepackage{booktabs}
\usepackage{textcomp}
\usepackage{lscape}
\usepackage{pifont}
\usepackage{algorithm}
\usepackage{graphicx}
\usepackage{caption}
\usepackage{subcaption}
\usepackage{CJKutf8}
\usepackage{algorithm}
\usepackage{algpseudocode}
\usepackage{enumitem}
\usepackage{wrapfig}
\usepackage{tabularx}
\usepackage{booktabs}
\usepackage{array}
\usepackage{pifont}
\usepackage{xcolor}
\usepackage{tabularray}

\usepackage{xcolor}

\title{Towards Scalable Customization and Deployment of Multi-Agent Systems for Enterprise Applications}

\author{Paresh Dashore$^\dagger$\thanks{Co-first authors. $^\dagger$Corresponding author. Email: \texttt{paresh.dashore@capitalone.com}.}, Shreyas Kulkarni$^*$, Uttam Gurram$^*$, Nadia Bathaee,\\ \textbf{Kartik Balasubramaniam, Genta Indra Winata, Sambit Sahu, Shi-Xiong Zhang}\\
AI Foundations, Capital One}
  
\begin{document}
\maketitle

\begin{abstract} 
% Large language model (LLM)-based multi-agent systems have shown strong capabilities across complex reasoning and task-completion tasks, creating significant opportunities for enterprise and industry applications. However, deploying these systems in production remains challenging due to the need for extensive customization to domain-specific requirements, as well as the high latency and inference costs associated with large-scale agentic workflows. In this work, we present a unified framework for both customization and efficient deployment of multi-agent systems in real-world enterprise settings. For the first stage, Agentic Model Customization, our approach combines continual pre-training, supervised fine-tuning, and preference optimization to adapt a compact dense model to specialized domains while preserving the capabilities of larger agentic systems. For the second stage, Inference Optimization, to enable cost-effective serving at scale, we further introduce a pipeline that integrates speculative decoding, FP8 quantization, and targeted calibration and training techniques, achieving substantial acceleration with minimal quality degradation. Across diverse enterprise workloads, our framework supports rapid adaptation to domain-specific requirements and delivers 4.48$\times$ end-to-end latency while maintaining task performance and improving robustness on long-tail scenarios.
Large language model (LLM)-based multi-agent systems demonstrate strong performance on complex reasoning and task execution, enabling broad enterprise applications. However, production deployment remains challenging due to domain-specific customization requirements and high latency and inference costs in agentic workflows. We propose a unified framework for customization and efficient deployment of multi-agent systems in real-world settings. The first stage, Agentic Model Customization, combines continual pretraining, supervised fine-tuning, and preference optimization to adapt a compact model to specialized domains while retaining strong agentic capabilities. The second stage, Inference Optimization, integrates speculative decoding and FP8 quantization with targeted calibration to enable cost-efficient serving with minimal quality loss. Across enterprise workloads, our framework enables rapid domain adaptation and achieves a 4.48$\times$ speedup in throughput while maintaining performance and improving robustness on long-tail scenarios.
\end{abstract}

\section{Introduction}
% The progress of Large Language Models (LLMs) has demonstrated their ability to address complex problems through agentic applications, such as tool-calling~\cite{chakraborty2026t1,shi2025tool,xu2025alignment} and multi-agent systems~\cite{guo2024large,wu2024autogen}. In multi-agent LLM applications, complex tasks are decomposed across specialized agents, often yielding higher-quality outputs than single-agent approaches. However, orchestrating multiple LLM calls introduces substantial latency and cost overheads. In production environments, this can be impractical and may fail to meet service-level agreement (SLA) requirements. Moreover, such systems typically rely on several large model instances, increasing infrastructure and operational expenses. As a result, workflows that require multiple LLM calls per request accumulate delays and costs, limiting their suitability for latency-sensitive, high-volume applications. Consequently, complex multi-agent systems often face deployment challenges, as high latency can negatively affect both performance and user experience.
The progress of Large Language Models (LLMs) enables agentic applications, including tool-calling~\cite{shi2025tool,xu2025alignment,chakraborty2026t1,winata2026t1} and multi-agent systems~\cite{guo2024large,wu2024autogen}. By decomposing complex tasks across specialized agents, multi-agent systems often achieve higher-quality outputs than single-agent approaches. However, coordinating multiple LLM calls incurs significant latency and computational overhead, making deployment challenging in production environments with strict service-level agreement (SLA) requirements. Moreover, the reliance on large models increases infrastructure costs and limits scalability for latency-sensitive, high-volume applications, potentially degrading the user experience.

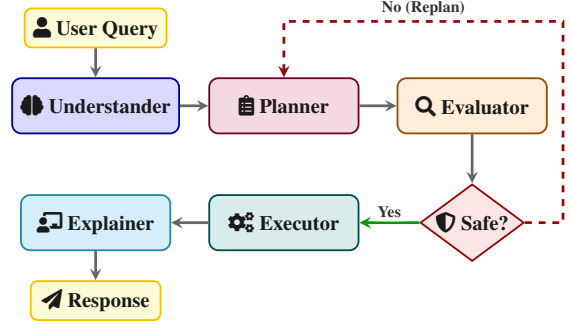
\begin{figure}[!t]
    \centering
    \resizebox{.95\columnwidth}{!}{%
        \begin{tikzpicture}[
            >=stealth, 
            font=\sffamily,
            %
            % ── Base style for all internal agents ──────────────────
            agent/.style={
                rectangle, rounded corners=4pt,
                minimum width=2.3cm, minimum height=0.85cm,
                align=center, font=\small\bfseries, thick, text=black!90
            },
            %
            % ── Fun, Light Pastel Colors ────────────────────────────
            ua/.style={agent, fill=blue!15, draw=blue!60!black},
            pa/.style={agent, fill=purple!15, draw=purple!60!black},
            ea/.style={agent, fill=orange!15, draw=orange!60!black},
            exa/.style={agent, fill=teal!15, draw=teal!60!black},
            xa/.style={agent, fill=cyan!15, draw=cyan!60!black},
            %
            % ── Separate Decision Diamond ───────────────────────────
            dec/.style={
                diamond, aspect=1.3,
                minimum width=1.6cm, minimum height=1.0cm,
                align=center, font=\footnotesize\bfseries,
                fill=red!10, draw=red!60!black, thick, text=black!90, inner sep=0pt
            },
            %
            % ── User Input/Output Nodes ─────────────────────────────
            io/.style={
                rectangle, rounded corners=3pt,
                minimum width=2.0cm, minimum height=0.6cm,
                align=center, font=\footnotesize\bfseries,
                fill=yellow!20, draw=orange!50!yellow, thick, text=black!90
            },
            %
            % ── Arrow & Label Styles ────────────────────────────────
            flow/.style={->, line width=1.2pt, draw=black!60},
            flowyes/.style={->, line width=1.2pt, draw=green!60!black},
            flowno/.style={->, line width=1.2pt, draw=red!60!black, dashed},
            lbl/.style={
                fill=white, inner sep=2pt, font=\scriptsize\bfseries,
                text=black!80, rounded corners=2pt
            }
        ]

        % ════════════════════════════════════════════════════════════
        %  COMPACT GRID PLACEMENT
        % ════════════════════════════════════════════════════════════
        
        % Top Row
        \node[io]  (user) at (0, 1.2)   {\faUser~User Query};
        \node[ua]  (ua)   at (0, 0)     {\faBrain~Understander};
        \node[pa]  (pa)   at (2.9, 0)   {\faClipboardList~Planner};
        \node[ea]  (ea)   at (5.8, 0)   {\faSearch~Evaluator};

        % Bottom Row (S-Curve)
        \node[dec] (dec)  at (5.8, -1.8) {\faShield*~Safe?};
        \node[exa] (exa)  at (2.9, -1.8) {\faCogs~Executor};
        \node[xa]  (xa)   at (0, -1.8)   {\faChalkboardTeacher~Explainer};
        \node[io]  (out)  at (0, -3.0)   {\faPaperPlane~Response};

        % ════════════════════════════════════════════════════════════
        %  CONNECTIONS
        % ════════════════════════════════════════════════════════════
        
        % Forward Pass
        \draw[flow] (user) -- (ua);
        \draw[flow] (ua) -- (pa);
        \draw[flow] (pa) -- (ea);
        
        % Evaluator to Decision
        \draw[flow] (ea) -- (dec);

        % Decision (Yes) -> ExA -> XA -> Output
        \draw[flowyes] (dec) -- node[lbl, above] {Yes} (exa);
        \draw[flow] (exa) -- (xa);
        \draw[flow] (xa) -- (out);

        % Decision (No) -> Feedback Loop to Planner
        % Routes out the right side, loops up around the Evaluator, and drops into the top of the Planner
        \draw[flowno] (dec.east) -- (7.2, -1.8) -- (7.2, 1.3) -- node[lbl, above] {No (Replan)} (2.9, 1.3) -- (pa.north);

        \end{tikzpicture}%
    }
    \caption{\textbf{Multi-Agent System Pipeline}. The sequential workflow routes a user query through specialized agents to produce a tool-based plan. Safety guardrails in the \textsc{Evaluator Agent} ensure that only valid plans proceed to execution and explanation, while invalid or unsafe plans trigger a replanning loop.}
    \label{fig:detailed_agent_workflow}
\end{figure}

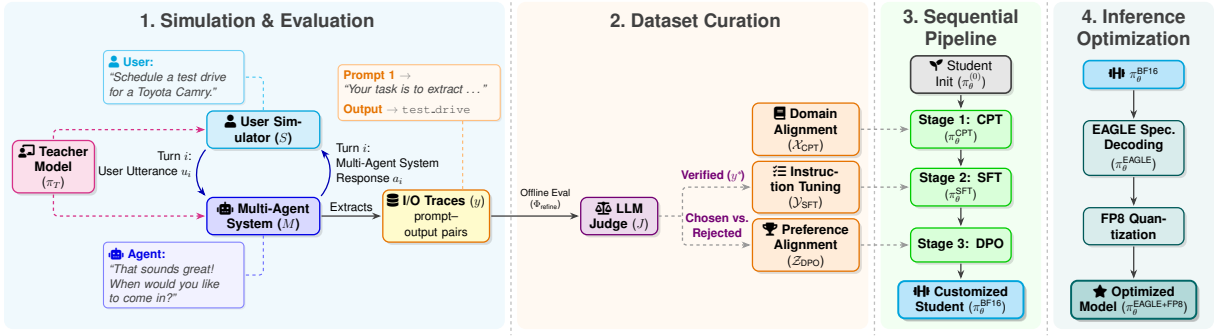
\begin{figure*}[t]
    \centering
    \resizebox{\textwidth}{!}{%
        \begin{tikzpicture}[
            >=Stealth,
            % ── Base style ──────────────────────────────────────────
            base_node/.style={
                draw, line width=1.2pt, rounded corners=6pt, align=center,
                font=\large\sffamily,
                drop shadow={opacity=0.25, shadow xshift=2.5pt, shadow yshift=-2.5pt}
            },
            % ── Content node styles ─────────────────────────────────
            sim/.style={base_node, fill=cyan!15, draw=cyan!90!black,
                        text width=3.4cm, minimum height=1.1cm},
            mas/.style={base_node, fill=blue!15, draw=blue!90!black,
                        text width=3.4cm, minimum height=1.1cm},
            teacher/.style={base_node, fill=magenta!15, draw=magenta!90!black,
                            text width=2.4cm, minimum height=1.0cm},
            judge/.style={base_node, fill=violet!15, draw=violet!90!black,
                          text width=2.238cm, minimum height=1.1cm},
            data/.style={base_node, fill=orange!15, draw=orange!90!black,
                         text width=3.2cm, minimum height=1.2cm},
            proc/.style={base_node, fill=green!15, draw=green!80!black,
                         text width=3.0cm, minimum height=0.9cm},
            proc_ckpt/.style={base_node, fill=green!15, draw=green!80!black,
                              text width=3.0cm, minimum height=1.1cm},
            ckpt/.style={base_node, fill=gray!15, draw=gray!70,
                         text width=2.2cm, minimum height=0.85cm,
                         font=\small\sffamily, drop shadow={opacity=0.15}},
            init_node/.style={base_node, fill=gray!20, draw=darkgray,
                              text width=3.0cm, minimum height=0.9cm,
                              line width=1.5pt},
            final_model/.style={base_node, fill=cyan!25, draw=cyan!90!black,
                                text width=3.8cm, minimum height=0.9cm,
                                line width=1.8pt, drop shadow={opacity=0.3, shadow xshift=3pt, shadow yshift=-3pt}},
            % ── Callout & Label styles ──────────────────────────────
            callout/.style={rounded corners=4pt, text width=4.2cm,
                            align=left, font=\normalsize\sffamily, inner sep=5pt,
                            drop shadow={opacity=0.12, shadow xshift=1.5pt, shadow yshift=-1.5pt}},
            judge_label/.style={font=\normalsize\sffamily\bfseries, text=violet!90!black, align=center},
            % ── Arrow styles ────────────────────────────────────────
            arrow/.style={->, thick, line width=1.2pt, draw=darkgray, shorten >=1pt, shorten <=1pt},
            loop_arrow/.style={->, thick, line width=1.2pt, draw=blue!70!black, shorten >=1pt, shorten <=1pt},
            dashed_arrow/.style={->, dashed, thick, line width=1.2pt, draw=gray!80, shorten >=1pt, shorten <=1pt},
            teacher_arrow/.style={->, dashed, thick, line width=1.2pt, draw=magenta!90!black, shorten >=1pt, shorten <=1pt},
            hdr/.style={font=\LARGE\bfseries\sffamily, text=darkgray, align=center},
            % ── Column 4 (Inference Optimization) styles ────────────
            infer_init/.style={base_node, fill=cyan!25, draw=cyan!90!black,
                               text width=3.0cm, minimum height=0.9cm, line width=1.5pt},
            infer_proc/.style={base_node, fill=teal!15, draw=teal!70!black,
                               text width=3.0cm, minimum height=1.2cm},
            infer_final/.style={base_node, fill=teal!30, draw=teal!80!black,
                                text width=3.8cm, minimum height=0.9cm,
                                line width=1.8pt, drop shadow={opacity=0.3, shadow xshift=3pt, shadow yshift=-3pt}}
        ]

        % ════════════════════════════════════════════════════════════
        %  BACKGROUND PANELS
        % ════════════════════════════════════════════════════════════
        \fill[cyan!5,   rounded corners=8pt] (-7.4, 5.8) rectangle ( 8.824, -4.8);
        \fill[orange!5, rounded corners=8pt] ( 9.224, 5.8) rectangle (20.6, -4.8);
        \fill[green!5,  rounded corners=8pt] (21.0, 5.8) rectangle (26.2, -4.8);
        \fill[teal!6,   rounded corners=8pt] (26.6, 5.8) rectangle (31.8, -4.8);

        % Dotted lines
        \draw[dashed, gray!50, line width=1.2pt] ( 9.024,  6.0) -- ( 9.024, -5.0);
        \draw[dashed, gray!50, line width=1.2pt] (20.8,  6.0) -- (20.8, -5.0);
        \draw[dashed, gray!50, line width=1.2pt] (26.4,  6.0) -- (26.4, -5.0);

        % ════════════════════════════════════════════════════════════
        %  SECTION HEADERS
        % ════════════════════════════════════════════════════════════
        \node[hdr] at ( 0.80, 5.2) {1.\ Simulation \& Evaluation};
        \node[hdr] at (15.00, 5.2) {2.\ Dataset Curation};
        \node[hdr] at (23.60, 4.919) {3.\ Sequential\\Pipeline};
        \node[hdr] at (29.20, 4.919) {4.\ Inference\\Optimization};

        % ════════════════════════════════════════════════════════════
        %  PART 1 — SIMULATION
        % ════════════════════════════════════════════════════════════
        \node[sim] (sim) at (1.0,  1.661) {\faUser\ \textbf{User Simulator ($S$)}};
        \node[mas] (mas) at (1.0, -1.139) {\faRobot\ \textbf{Multi-Agent System ($M$)}};

        % ── Teacher Model
        \node[teacher] (teacher) at (-5.8, 0.461) {\faChalkboardTeacher\ \textbf{Teacher Model}\\($\pi_T$)};
        \draw[teacher_arrow, rounded corners=4pt] (teacher.north) |- (sim.west);
        \draw[teacher_arrow, rounded corners=4pt] (teacher.south) |- (mas.west);

        % ── Enhanced Callout: user utterance
        \node[callout, draw=cyan!50, fill=cyan!5, line width=1pt] (ex_sim) at (-1.903, 3.361) {%
            \textcolor{cyan!90!black}{\textbf{\faUser\ User:}}\\
            \textcolor{darkgray}{\itshape ``Schedule a test drive\\for a Toyota Camry.''}};
        \draw[-, dashed, cyan!60, line width=1pt, rounded corners=4pt] (sim.north) |- (ex_sim.east);

        % ── Enhanced Callout: agent response
        \node[callout, draw=blue!50, fill=blue!5, line width=1pt] (ex_resp) at (-1.903, -3.050) {%
            \textcolor{blue!90!black}{\textbf{\faRobot\ Agent:}}\\
            \textcolor{darkgray}{\itshape ``That sounds great!\\When would you like\\to come in?''}};
        \draw[-, dashed, blue!60, line width=1pt, rounded corners=4pt] (mas.south) |- (ex_resp.east);

        % ── Conversational loop arrows
        \draw[loop_arrow] (sim.south west)
            to[bend right=40]
            node[left, font=\normalsize\sffamily, align=right, xshift=-1pt, yshift=8pt]
                {Turn $i$:\\ User Utterance $u_i$}
            (mas.north west);
        \draw[loop_arrow] (mas.north east)
            to[bend right=40]
            node[right, font=\normalsize\sffamily, align=left, xshift=2pt, yshift=8pt]
                {Turn $i$:\\ Multi-Agent System\\Response $a_i$}
            (sim.south east);

        % ── I/O Traces
        \node[data, fill=yellow!25, draw=orange!90!black] (raw) at (6.6, -1.139) {%
            \faDatabase\ \textbf{I/O Traces} ($y$)\\
            \normalsize prompt--output pairs};
        \draw[arrow] (mas.east)
            -- node[above, font=\normalsize\sffamily] {Extracts}
            (raw.west);

        % ── Enhanced Callout: I/O Traces Example
        \node[callout, draw=orange!50, fill=orange!5, line width=1pt, text width=4.9cm] (ex_raw) at (6.05, 2.861) {%
            \textcolor{orange!90!black}{\textbf{Prompt 1}} \textcolor{gray}{$\rightarrow$}\\
            \textcolor{darkgray}{\itshape ``Your task is to extract \dots''}\\[4pt]
            \textcolor{orange!90!black}{\textbf{Output}} \textcolor{gray}{$\rightarrow$} \textcolor{darkgray}{\ttfamily test\_drive}};
        \draw[-, dashed, orange!60, line width=1pt] ([xshift=0.859cm]raw.north) -- ([xshift=0.859cm]raw.north |- ex_raw.south);

        % ════════════════════════════════════════════════════════════
        %  PART 2 — DATASET CURATION
        % ════════════════════════════════════════════════════════════
        \node[judge] (judge) at (12.531, -1.139) {\faBalanceScale\ \textbf{LLM Judge} ($J$)};

        % Using pos=0.65 to drop the text securely between the divider and the Judge
        \draw[arrow] (raw.east)
            -- node[above, align=center, font=\small\sffamily, yshift=2pt, pos=0.579, xshift=3pt]
               {Offline Eval \\ ($\Phi_{\text{refine}}$)}
            (judge.west);

        \node[data] (cpt_data) at (18.576,  1.673) {%
            \faBook\ \textbf{Domain Alignment}\\($\mathcal{X}_{\text{CPT}}$)};
        \node[data] (sft_data) at (18.576, -0.202) {%
            \faTasks\ \textbf{Instruction Tuning}\\($\mathcal{Y}_{\text{SFT}}$)};
        \node[data] (dpo_data) at (18.576, -2.076) {%
            \faTrophy\ \textbf{Preference Alignment}\\($\mathcal{Z}_{\text{DPO}}$)};

        % ── Orthogonal Routing for Judge Output
        \draw[dashed, thick, line width=1.2pt, draw=gray!80] (judge.east) -- (14.580, -1.139);

        \draw[dashed_arrow, rounded corners=4pt] (14.580, -1.8) |-
            node[above, pos=0.5, font=\normalsize\sffamily\bfseries, text=violet!90!black, yshift=1pt, xshift=28pt] {Verified ($y^*$\!)}
            (sft_data.west);

        \draw[dashed_arrow, rounded corners=4pt] (14.580, -1.8) |-
            node[above, pos=0.5, align=center, font=\normalsize\sffamily\bfseries, text=violet!90!black, yshift=1pt, xshift=32pt] {Chosen vs.\\Rejected}
            (dpo_data.west);

        % ════════════════════════════════════════════════════════════
        %  PART 3 — SEQUENTIAL PIPELINE
        % ════════════════════════════════════════════════════════════
        \node[init_node]    (init)   at (23.6,  3.447)  {\faSeedling\ Student Init ($\pi_{\theta}^{(0)}$)};

        \node[proc_ckpt]    (cpt)    at (23.6,  1.673)  {\textbf{Stage 1: CPT}\\[1pt]\normalsize($\pi_{\theta}^{\text{CPT}}$)};

        \node[proc_ckpt]    (sft)    at (23.6, -0.202)  {\textbf{Stage 2: SFT}\\[1pt]\normalsize($\pi_{\theta}^{\text{SFT}}$)};

        \node[proc_ckpt]    (dpo)    at (23.6, -2.076)  {\textbf{Stage 3: DPO}};

        \node[final_model]  (final)  at (23.6, -3.850)  {%
            \faDumbbell\ \textbf{Customized Student} ($\pi_{\theta}^{\text{BF16}}$)};

        % Waterfall flow
        \draw[arrow] (init.south) -- (cpt.north);
        \draw[arrow] (cpt.south)  -- (sft.north);
        \draw[arrow] (sft.south)  -- (dpo.north);
        \draw[arrow] (dpo.south)  -- (final.north);

        % Dataset feeds into training stages
        \draw[dashed_arrow] (cpt_data.east) -- (cpt.west);
        \draw[dashed_arrow] (sft_data.east) -- (sft.west);
        \draw[dashed_arrow] (dpo_data.east) -- (dpo.west);

        % ════════════════════════════════════════════════════════════
        %  PART 4 — INFERENCE OPTIMIZATION (column 4)
        % ════════════════════════════════════════════════════════════
        \node[infer_init]   (init2)   at (29.2,  3.447)  {\faDumbbell\ $\pi_{\theta}^{\text{BF16}}$};

        \node[infer_proc]   (eagle2)  at (29.2,  1.065)  {\textbf{EAGLE Spec. Decoding}\\[3pt]($\pi_{\theta}^{\text{EAGLE}}$)};

        \node[infer_proc]   (fp82)    at (29.2, -1.468)  {\textbf{FP8 Quantization}};

        \node[infer_final]  (final2)  at (29.2, -3.850)  {%
            \faStar\ \textbf{Optimized Model} ($\pi_{\theta}^{\text{EAGLE+FP8}}$)};

        % Waterfall flow
        \draw[arrow] (init2.south)  -- (eagle2.north);
        \draw[arrow] (eagle2.south) -- (fp82.north);
        \draw[arrow] (fp82.south)   -- (final2.north);

        \end{tikzpicture}%
    }
    \caption{%
        \textbf{End-to-End Pipeline.} An end-to-end flow distilling agentic capabilities from a teacher model ($\pi_T$) into a customized student ($\pi_\theta^{\text{BF16}}$) and further optimizing it through inference techniques into $\pi_{\theta}^{\text{EAGLE+FP8}}$ model.
    }
    \label{fig:agentic_finetuning}
\end{figure*}

At the same time, deployed agentic systems must retain strong task-specific capabilities. Skill transfer in LLMs emerges as a key approach to enabling dense models to acquire multiple competencies through fine-tuning~\cite{nottingham2024skill, wang2025re}. This is particularly important in agentic settings, where a single model is expected to perform specialized roles without relying on multiple independent models that are harder to maintain and scale. In addition, compressing knowledge into smaller models, commonly achieved through model distillation, is essential for accelerating inference while preserving performance comparable to larger models. Meanwhile, speculative decoding~\cite{leviathan2023fast,li2024eagle} proves to be an effective technique for reducing latency by leveraging smaller models during inference.

To address the deployment challenges of multi-agent systems, we propose an agentic model customization and inference optimization pipeline that substantially reduces latency while preserving strong task performance. Our approach begins with model distillation using a student–teacher framework to consolidate agentic capabilities into a single optimized model. The pipeline further leverages unlabeled data for domain adaptation and knowledge transfer via Continual Pretraining (CPT), incorporates supervised fine-tuning (SFT) during distillation, and applies post-training Direct Preference Optimization (DPO) \citep{rafailov2023direct} to better align model behavior with desired preferences. Finally, we enhance inference efficiency through a combination of EAGLE speculative decoding~\cite{li2024eagle} and FP8 quantization, achieving additional latency reductions with minimal impact on model quality. We illustrate our sequential multi-agent pipeline in Figure~\ref{fig:detailed_agent_workflow}.

Our contributions are summarized as follows:
\begin{itemize}
    \item We propose a production-ready multi-agent system that integrates both \textbf{agentic model customization} and an \textbf{inference optimization} pipeline for real-world enterprise deployments. The former distills agentic capabilities into a smaller model, whereas the latter preserves model performance while significantly reducing inference latency via EAGLE speculative decoding and FP8 quantization.
    % \item An inference optimization pipeline including FP8 quantization and EAGLE speculative decoding, that substantially reduce latency while preserving accuracy on real-world agentic workloads, allowing sub-second tail latency. 
    \item We present an end-to-end (E2E) training pipeline comprising a user-simulator-driven data generation framework and a sequential training process for customized agentic models. Through a systematic analysis of each training stage, we quantify its contribution to production-grade quality and show that preference optimization is essential for achieving competitive performance.
    \item We conduct comprehensive empirical studies demonstrating that carefully curated mixtures of proprietary and public data enable near-lossless acceleration, EAGLE can be tuned to reach optimal efficiency and lower latency even while speculating less correct tokens.
\end{itemize}
%\section{Preference-Driven Agentic Fine-Tuning}
\section{Agentic Model Customization}
\label{sec:distillation}

Our system comprises a customer-facing chatbot for automotive retail, governed by an LLM-powered multi-agent workflow. To reduce operational latency, we implement an offline distillation and optimization pipeline to transition from a high-parameter production \textit{teacher model}, $\pi_T$, to an optimized \textit{student model}, $\pi_{\theta}$. All training stages within this customization phase are conducted in \texttt{BF16} precision.

\subsection{Multi-Agent System Architecture}
\label{sec:system_architecture}

To support complex customer interactions in automotive retail, we develop a Multi-Agent System ($M$) powering a customer-facing chatbot. All agents share the same foundation model but differ in context, including memory, knowledge bases, and tool access. This single-model design simplifies production deployment while preserving agent specialization.

As illustrated in Figure~\ref{fig:detailed_agent_workflow}, the system $M$ follows a sequential pipeline of five agents with a planning feedback loop. The system decomposes complex queries across specialized, collaborative roles: the \textsc{Understander Agent}, \textsc{Planner Agent}, \textsc{Evaluator Agent},  \textsc{Executor Agent}, and \textsc{Explainer Agent}. A comprehensive breakdown of each agent's specific responsibilities is provided in Appendix~\ref{sec:appendix_agent_roles}.

Because a single user request may require multiple multi-turn exchanges and replanning iterations, the cumulative latency and compute costs escalate quickly. Our goal is to maximize throughput on AWS EC2 P5 (8$\times$ NVIDIA H100 80GB GPUs) while meeting sub-second end-to-end latency SLAs. However, profiling identifies three primary bottlenecks: 
(1) cumulative latency from multiple LLM calls per request; 
(2) massive memory footprints from serving large LLMs; and 
(3) high generation costs that cannot be solved by prefill optimization alone. This compounding inference overhead necessitates the aggressive distillation and inference optimization strategies detailed in the subsequent sections. Further details regarding our specific deployment constraints and system profiling can be found in Appendix~\ref{sec:appendix_deployment_constraints}.

\subsection{Conversational Data Synthesis via Agentic Simulation}
\label{subsec:agentic_simulation}
To curate a high-fidelity training corpus, we develop an automated user simulation framework where a specialized User Simulator ($U$) models human customer interactions. As illustrated in the end-to-end pipeline in Figure~\ref{fig:agentic_finetuning}, the simulator is driven by an optimized prompt configuration, $\Phi_{S}$, engineered to maximize conversational diversity and expose the system to complex edge cases. Specifically, the simulation prompt $\Phi_{S}$ dynamically ingests four distinct context vectors at each turn: (i) the accumulated conversation history $H$, (ii) a targeted set of intent and capability definitions $\mathcal{N}$ mapping to supported business logic, (iii) seed topics $\mathcal{I}$ used to anchor the initial dialogue domain, and (iv) environmental context $\mathcal{E}$, which encompasses available vehicle inventory constraints and synthetic customer profiles.

A single simulation session $T$ continues until the simulator achieves its assigned goal and outputs an \texttt{EXIT} token. The interaction follows a sequential turn-taking logic. For each exchange $i$, the simulator generates a user utterance $u_i$ conditioned on the prior history $H_{<i}$, and the multi-agent system $M$ (powered by the teacher model $\pi_T$) generates an assistant response $a_i$:
\begin{align}
u_i &= S(H_{<i}, \mathcal{N}, \mathcal{I}, \mathcal{E}, \Phi_{S}), \\
a_i &= M(u_i \mid \pi_T).
\end{align}
During simulation, we capture the complete internal state, including all intermediate LLM prompts and corresponding outputs.

\subsection{Refinement via LLM-as-a-Judge}
To ensure the distillation of high-quality reasoning, we utilize a model acting as a judge, $J$. We define $J$ such that its reasoning capabilities and parameter scale significantly exceed the teacher model ($J \gg \pi_T$). For every teacher-generated response $y \in T$, the judge generates a refined response $y^*$ using a specialized instruction-adherence prompt $\Phi_\text{refine}$:
\begin{equation}
y^* = J(y, \Phi_\text{refine}).
\end{equation}
\subsection{Dataset Formulation}
Using the refined traces, we construct three corpora for the student model: (i) \textbf{Domain Alignment ($\mathcal{X}_{\text{CPT}}$)}, comprising synthetic and public unlabeled datasets, alongside domain-specific automotive texts; (ii) \textbf{Instruction Tuning ($\mathcal{Y}_{\text{SFT}}$)}, containing judge-refined outputs $y^*$ to distill teacher proficiency; and (iii) \textbf{Preference Alignment ($\mathcal{Z}_{\text{DPO}}$)}, consisting of chosen/rejected triples $(x, y^*, y)$. 
% alongside 1,000 hard negative where both models fail business-logic checks.

\subsection{Agentic Training Procedure}\label{ssec:training_procedure}

\paragraph{Stage 0: Model Curation.}

We initialize the policy model $\pi_{\theta}^{(0)}$ by applying block expansion \cite{wu2024llama} to a base foundation model $\pi_{\text{base}}$ \cite{grattafiori2024llama}. Specifically, we insert one new transformer block after every four original blocks to increase model capacity for domain-specific adaptation. The attention and feed-forward weights of each inserted block are initialized to zero. Because of the residual connections, these blocks initially act as identity mappings, allowing hidden states to pass through unchanged. Consequently, the expanded model $\pi_{\theta}^{(0)}$ retains the exact behavior and performance of $\pi_{\text{base}}$ at initialization, consistent with the findings of \citet{wu2024llama}.

% We adopt a sequential training regime where parameters are merged back into the foundational weights at the conclusion of every stage.

% \paragraph{Stage 1: Context-aware Continual Pretraining.}
% Previously work rely on continual pretraining to adapt existing pretrained models to new data. As the model's parameters are updated to assimilate new information, it can abruptly lose proficiency on previously learned domains, a phenomenon known as catastrophic forgetting \cite{vandeven2024continuallearningcatastrophicforgetting}. To address this issue, we propose Context-aware Continual Pretraining, a simple technique that provides the model with sample-specific context before adapting its weights to new content in order to smoothen the training loss. 
% \begin{equation}
% \mathcal{L}_{\text{CPT}}(\theta) = -\mathbb{E}_{x} \Bigl[ \sum_{i} \log P_{\theta}(x_t |  x_{<t}; C_x)  \Bigr],
% \end{equation}
% where $C_x$ is the context generated for each training document. The creation of $C_x$ is out of the scope of this paper.
% To further mitigate the forgetting, we performance model merging after every training stage.

\paragraph{Stage 1: Context-aware Continual Pretraining.}
Continual pretraining is widely used to adapt pretrained models to new domain data, but updating model parameters on new distributions can degrade previously acquired capabilities, a phenomenon known as catastrophic forgetting \cite{winata2023overcoming}. Our method is motivated by the observation that the per-token loss is consistently much higher at the beginning of each training sequence, as shown in Figure~\ref{fig:loss_token_pos}. In the CPT setting, where the model already possesses strong general linguistic knowledge, this initial loss spike is often caused by limited preceding context rather than a true failure to model the domain content, making it an inefficient and potentially noisy training signal. To reduce this effect, we propose \emph{Context-aware Continual Pretraining}, which prepends sample-specific context $C_x$ to each training document before updating the model, thereby smoothing the token-level loss and reducing abrupt distributional shifts during adaptation. The full mathematical formulation of the loss ($\mathcal{L}_{\tt{CA}-\tt{CPT}}$) is detailed in \Cref{sec:appendix_math_formulations}. To further mitigate forgetting, we perform model merging after each training stage to combine domain-specific adaptation with the general capabilities of the original model, yielding the merged model $\pi_{\theta}^{\tt{CPT}}$.

\paragraph{Stage 2: Agentic Fine Tuning.}
Starting from the merged $\pi_{\theta}^{\tt{CPT}}$, we perform Supervised Fine-Tuning on $\mathcal{Y}_{\text{SFT}}$ using Low-Rank Adaptation (LoRA). We avoid full-parameter fine-tuning to prevent catastrophic forgetting and ensure robustness to future prompt updates. The adapters are merged to create $\pi_{\theta}^{\tt{SFT}}$. The SFT loss objective ($\mathcal{L}_{\tt{SFT}}$) is provided in \Cref{sec:appendix_math_formulations}.

\paragraph{Stage 3: Agentic Preference Tuning.}
Using $\pi_{\theta}^{\tt{SFT}}$ as the reference, we apply DPO~\citep{rafailov2023direct} using LoRA on $\mathcal{Z}_{\text{DPO}}$. This stage aligns the student with the judge's logic and corrects teacher errors. The complete optimization objective ($\mathcal{L}_{\tt{DPO}}$) is explicitly outlined in \Cref{sec:appendix_math_formulations}. The final adapters are merged to yield the optimized student model, $\pi_{\theta}^{\texttt{BF16}}$, acting as the foundational \texttt{BF16} checkpoint for downstream inference optimization.

% Speculative decoding accelerates generation in a lossless way by using a draft model to predict tokens that are verified in parallel by target model~\cite{leviathan2023fast,chen2023accelerating}. EAGLE~\cite{li2024eagle,li2024eagle2,li2025eagle3} trains lightweight draft adapters from target-model hidden states to increase token acceptance rates. Prior work has limited analysis on how training data affects acceptance rates in domain-specific applications. We show that training data alignment materially affects acceptance rates, independent of the decoding algorithm.

% Specifically, we compare the performance of three adapters trained using: (1)~\textbf{External (E)}: ShareGPT (77k samples); (2)~\textbf{Synthetic (S)}: synthetic data generated from $M$ by simulating multi-turn conversations with responses from $\pi_{\theta}^{\texttt{BF16}}$ (50k samples); and (3)~\textbf{Combined (C)}: 77k external + 50k synthetic samples. We denote the EAGLE-augmented student as $\pi_{\theta}^{\texttt{EAGLE}}$

\section{Inference Optimization}
\label{sec:infopt}
\subsection{EAGLE}
\label{sec:eaglehere}
Speculative decoding utilizes a draft model to predict tokens that are verified in parallel by the target model~\cite{chen2023accelerating,leviathan2023fast}. Crucially, it accelerates generation while guaranteeing to preserve accuracy. EAGLE~\cite{li2024eagle,li2024eagle2,li2026eagle} is a lightweight draft model that consumes target-model hidden states to increase token acceptance rates, yielding higher throughput. However, prior work has limited analysis of training the EAGLE drafter for domain-specific applications. We demonstrate that, with carefully curated training data, we can achieve significant throughput improvements, independent of the decoding algorithm. We denote the EAGLE-augmented student as $\pi_{\theta}^{\texttt{EAGLE}}$.
We detail the architectural trade-offs of draft model quantization and tree versus greedy decoding in Appendix~\ref{sec:appendix_takeaways_model_quantization}.

\subsection{FP8 Post-Training Quantization}
\label{sec:quantization}

% \paragraph{Setup.}
FP8 quantization \cite{kuzmin2022fp8} compresses weights and activations resulting in lower memory requirements and faster computations leading to a reduction in latency. Additionally, quantizing the KV cache to FP8 halves the storage compared to FP16/BF16, effectively allowing for doubled context lengths or larger batch sizes. We apply FP8 (E4M3) weight-and-activation quantization (W8A8) to the $\pi_{\theta}^{\texttt{BF16}}$  (\Cref{sec:distillation}) and $\pi_{\theta}^{\texttt{EAGLE}}$ models using min-max per-tensor post-training quantization (PTQ), yielding the quantized student $\pi_{\theta}^{\texttt{FP8}}$ and optimized model $\pi_{\theta}^{\texttt{EAGLE+FP8}}$, respectively. Static quantization sets scales offline from activation statistics collected on calibration data, avoiding runtime calibration overhead. We select W8A8 over weight-only schemes (e.g., AWQ, GPTQ) because W8A8 maintains throughput gains under high concurrency when inference becomes compute-bound, and the FP8 format minimizes accuracy regression relative to integer quantization.
% Similar to EAGLE, we evaluate three calibration datasets against a baseline: a public calibration set (P) containing 1.4k CNN/DailyMail samples, an in-domain calibration set (I) containing 5.8k synthetic traces, and a mixed calibration set (M) combining both sources.
% \subsection{EAGLE}
% \label{sec:eaglehere}
% Speculative decoding accelerates generation in a lossless way by using a draft model to predict tokens that are verified in parallel by target model~\cite{leviathan2023fast,chen2023accelerating}. EAGLE~\cite{li2024eagle,li2024eagle2,li2025eagle3} trains lightweight draft adapters from target-model hidden states to increase token acceptance rates. Prior work has limited analysis of how training data affects acceptance rates in domain-specific applications. We show that training data alignment materially affects acceptance rates, independent of the decoding algorithm.
% Specifically, we compare the performance of three adapters trained using: (1)~\textbf{External (E)}: ShareGPT (77k samples); (2)~\textbf{Synthetic (S)}: synthetic prompts from $M$ generated by simulating multi-turn conversations with an LLM (50k samples); and (3)~\textbf{Combined (C)}: 77k external + 50k synthetic samples. We denote the EAGLE-augmented student as $\pi_{\theta}^{\texttt{EAGLE}}$, and the model that stacks FP8 quantization with EAGLE speculative decoding as $\pi_{\theta}^{\texttt{FP8+EAGLE}}$.
The optimization methods described in Sections~\ref{sec:eaglehere} and~\ref{sec:quantization} are complementary and can be stacked on top of the $\pi_{\theta}^{\texttt{BF16}}$ model to produce compounding gains as shown in Table~\ref{tab:quantization-eagle}.

\section{Experimental Setup}
\label{sec:experimental_setup}

We construct a customized 10B model, denoted by $\pi_{\theta}^{(0)}$, following the Stage 0 procedure. This model is built upon Llama 3.1 8B Instruct~\cite{grattafiori2024llama}, which serves as the base policy model $\pi_{\text{base}}$. The CPT stage consumes approximately 5T tokens, consisting of a mixture of in-domain and public-domain data. During the agentic fine-tuning (AFT) stage, we use a larger teacher model, $\pi_\text{T}$ with Llama 3 70B Instruct, to curate the supervised fine-tuning corpus. We train a 250M EAGLE drafter using responses and hidden states generated by $\pi_{\theta}^{\texttt{BF16}}$ on a combined dataset of 127k samples, consisting of 77k open-source dialogue traces and 50k proprietary synthetic simulations from $M$.
% We generate responses and hidden states using the $\pi_{\theta}^{\texttt{BF16}}$ model to train a 250M EAGLE drafter on top of it. For this experiment, we train three EAGLE drafters using on (1) \textbf{external} ShareGPT data (77k samples), (2) \textbf{synthetic} data generated from $M$ via multi-turn simulations with responses from $\pi_{\theta}^{\texttt{BF16}}$ (50k samples), and (3) a \textbf{combined} dataset (77k + 50k) with \texttt{BF16} precision.

% Following the training phase, we utilize AWS EC2 P5 to run FP8 calibration experiments and measure inference performance. Similar to EAGLE, three calibration sets: (1) a \textbf{public} set containing 1.4k data from~\citet{nallapati2016abstractive}, (2) an \textbf{in-domain} set containing 5.8k synthetic traces, and (3) a \textbf{mixed} set combining both sources. 
Following the training phase, we utilize AWS EC2 P5 instances to measure inference performance and run FP8 calibration experiments on a mixed dataset, comprising 1.4k public samples from~\citet{nallapati2016abstractive} and 5.8k in-domain synthetic traces from $M$. All inference metrics are obtained using the NVIDIA TensorRT-LLM v19 framework. A comprehensive breakdown of our training infrastructure, detailed dataset synthesis, and sequential distillation hyperparameters is provided in Appendix~\ref{sec:appendix_training_details}.

\begin{table}[!t]
\centering
\setlength{\tabcolsep}{5pt}
\resizebox{.49\textwidth}{!}{%
\begin{tabular}{lc cccc}
\toprule
\textbf{Configuration} & \textbf{Decoding} & \textbf{MGL} & \textbf{Latency (s)} & \textbf{QPS} & \textbf{Speedup} \\
\midrule
Llama 3 70B ($\pi_T$) & --      & --  & 3.92 & 1.46 & 1.00$\times$  \\
  Baseline ($\pi_{\theta}^{\texttt{BF16}}$)    & --      & --  & 1.69 & 3.40 & 2.33$\times$  \\
  Baseline ($\pi_{\theta}^{\texttt{FP8}}$)   & --      & --  & 1.60 & 3.66 & 2.50$\times$    \\
  $\pi_{\theta}^{\texttt{EAGLE+FP8}}$   & Greedy  &   3.80  & 0.92 & 6.54 & 4.48$\times$    \\
\bottomrule
\end{tabular}
}
\caption{P90 latency and throughput across optimization configurations. MGL = Mean Generated Length; QPS = Queries Per Second. $\pi_{\theta}^{\texttt{EAGLE+FP8}}$ combines EAGLE with FP8 quantization. All configurations use BF16 unless noted otherwise.}
\label{tab:quantization-eagle}
\vspace{-1mm}
\end{table}

\begin{table}[!t]
\centering
\setlength{\tabcolsep}{5pt}
\resizebox{.49\textwidth}{!}{%
\begin{tabular}{lc cccc}
\toprule
\textbf{Configuration} & \textbf{Decoding} & \textbf{MGL} & \textbf{Latency (s)} & \textbf{QPS} & \textbf{Speedup} \\
\midrule
  Baseline ($\pi_{\theta}^{\texttt{BF16}}$)    & --      & --  & 1.69 & 3.40 & 1.00$\times$  \\
  $\pi_{\theta}^{\texttt{EAGLE}}$ (E)       & Tree    &  3.66  & 1.50 & 4.04 & 1.19$\times$    \\
  $\pi_{\theta}^{\texttt{EAGLE}}$ (S)        & Tree    &  3.98   & 1.29 & 4.66 & 1.37$\times$    \\
  $\pi_{\theta}^{\texttt{EAGLE}}$ (C)      & Tree    &   4.29  & 1.19 & 4.96 & 1.46$\times$    \\
  $\pi_{\theta}^{\texttt{EAGLE}}$ (S)      & Greedy    &   3.53 & 1.13 & 5.50 & 1.62$\times$    \\
  $\pi_{\theta}^{\texttt{EAGLE}}$ (C)      & Greedy    &   3.80  & 0.96 & 6.07 & 1.78$\times$    \\
\bottomrule
\end{tabular}
}
\caption{Performance across different EAGLE configurations. E = External, S = Synthetic, C = Combined.}
\label{tab:eagle-ablations}
\vspace{-3mm}
\end{table}

\begin{figure*}[!th]
    \centering
    \includegraphics[width=\textwidth]{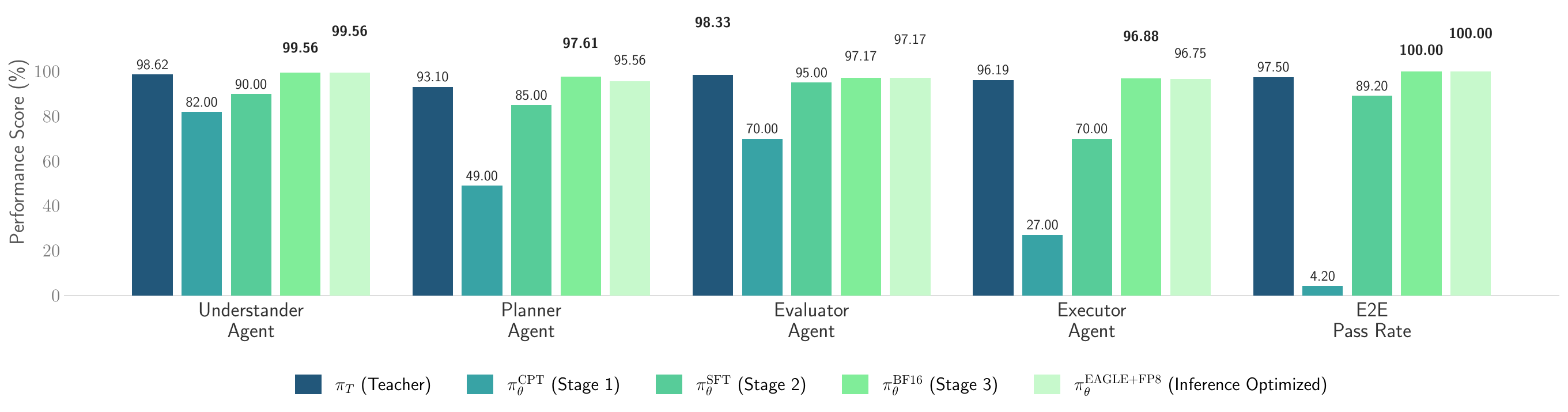}
    \caption{Performance evaluation of the multi-agent system across different agents and training stages, evaluated on a held-out set of 1,424 simulated conversations (8,848 data points). It also includes End-to-End functional stress test pass rates across 120 complex scenarios.}
    \label{fig:f1cc_results}
    % \vspace{-3mm}
\end{figure*}

\begin{figure}[!t]
\centering
\begin{tikzpicture}
\begin{axis}[
    width=\linewidth,
    height=5.5cm,
    xlabel={\small Calibration sequence length (tokens)},
    ylabel={\small Activation clip rate ($\times 10^{-5}$\%)},
    xmode=log,
    ymode=log,
    log basis x=2,
    log basis y=10,
    xtick={32,64,128,256,512,1024,2048},
    xticklabels={32,64,128,256,512,1024,2048},
    xticklabel style={font=\footnotesize},
    yticklabel style={font=\footnotesize},
    label style={font=\footnotesize},
    grid=both,
    grid style={gray!18, thin},
    major grid style={gray!28, thin},
    legend cell align={left},
    legend style={
        font=\tiny,
        at={(0.98,0.98)},
        anchor=north east,
        draw=gray!50,
        fill=white,
        fill opacity=0.92,
        text opacity=1,
        row sep=-1pt,
        inner sep=3pt,
        text width=1.4cm
    }
]

\addlegendimage{empty legend}
\addlegendentry{\hfill\shortstack[c]{\textbf{E2E}\\\textbf{Pass Rate}}}

% Public (P) — thin reference line
\addplot[
    color=blue!55!black,
    line width=0.9pt,
    densely dashed,
    mark=square*,
    mark size=2.4pt,
    mark options={solid, fill=blue!55!black, draw=blue!55!black}
] coordinates {
    (32,    0.7574)
    (64,    0.3982)
    (128,   0.1615)
    (256,   0.0467)
    (512,   0.0163)
    (1024,  0.0064)
    (2048,  0.0055)
};
\addlegendentry{P \hfill 97.27\%\hspace{0.2cm}}

% In-domain (I) — thin reference line
\addplot[
    color=orange!80!black,
    line width=0.9pt,
    densely dotted,
    mark=triangle*,
    mark size=2.8pt,
    mark options={solid, fill=orange!80!black, draw=orange!80!black}
] coordinates {
    (32,    4.8040)
    (64,    2.2633)
    (128,   0.6821)
    (256,   0.0912)
    (512,   0.0199)
    (1024,  0.0070)
    (2048,  0.0013)
};
\addlegendentry{I \hfill 98.18\%\hspace{0.2cm}}

% Mixed (M) — hero line: thicker, solid
\addplot[
    color=green!42!black,
    line width=1.8pt,
    mark=*,
    mark size=3.0pt,
    mark options={solid, fill=green!42!black, draw=green!42!black}
] coordinates {
    (32,    1.0405)
    (64,    0.5652)
    (128,   0.2184)
    (256,   0.0363)
    (512,   0.0089)
    (1024,  0.0032)
    (2048,  0.0009)
};
\addlegendentry{\textbf{M} \hfill \textbf{100.00\%}\hspace{0.2cm}}

\end{axis}
\end{tikzpicture}
\caption{Activation clip rate ($\times 10^{-5}$\%) vs.\ calibration sequence length for $\pi_{\theta}^{\texttt{BF16}}$. P = Public, I = In-domain (Synthetic), M = Mixed (P+I). 1024 calibration/test samples; 8192-token test length; both axes log-scaled.}
\label{fig:clip_rate}
% \vspace{-3mm}
\end{figure}
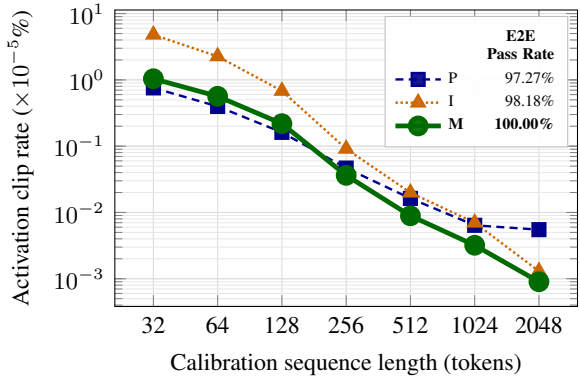

\section{Results and Analysis}
\label{sec:results-analysis}

We measure distillation success and inference optimizations using agent-level evaluations (1,424 simulated conversations, 8,848 data points) and an E2E functional stress test (120 scenarios, ~10 turns each) that simulates mid-conversation business switches. Figure~\ref{fig:f1cc_results} summarizes these results.

\paragraph{Distillation Progression.}
Task metrics validate our multi-stage pipeline (Figure~\ref{fig:f1cc_results}). Continual Pretraining alone ($\pi_\theta^{\text{CPT}}$) yields poor agentic capabilities and near-total E2E failure due to weak instruction-following. Supervised Fine-Tuning ($\pi_\theta^{\text{SFT}}$) establishes foundational task structures, driving substantial improvements and a higher E2E success rate. DPO bridges the remaining capability gap. The final distilled student ($\pi_{\theta}^{\texttt{BF16}}$) outperforms the 70B teacher ($\pi_T$) in \textsc{Planner Agent} and \textsc{Understander Agent} (Figure~\ref{fig:f1cc_results}), and navigates all E2E scenarios to exceed the teacher's baseline. Additionally, $\pi_{\theta}^{\texttt{BF16}}$ achieves a 2.33$\times$ speedup over $\pi_T$ (Table~\ref{tab:quantization-eagle}).

\paragraph{Inference Optimization Impact.}
Stacking optimization techniques produces $\pi_{\theta}^{\texttt{EAGLE+FP8}}$, which runs 1.92$\times$ faster than $\pi_{\theta}^{\texttt{BF16}}$ and 4.48$\times$ faster than $\pi_T$ (Table~\ref{tab:quantization-eagle}). Task-level accuracy remains highly resilient under quantization and speculative decoding: $\pi_{\theta}^{\texttt{EAGLE+FP8}}$ maintains near-identical performance to the unquantized student, with only a minor regression in the Planner agent that still exceeds the teacher baseline (Figure~\ref{fig:f1cc_results}). Crucially, $\pi_{\theta}^{\texttt{EAGLE+FP8}}$ retains a perfect E2E stress test pass rate, confirming our calibration and training data mixing strategies preserve agentic behavior.

\paragraph{EAGLE Alignment.}
\label{sec:eagle-perf}
We analyze the performance of three EAGLE drafters trained on the combined dataset and its individual external and synthetic subsets. Table~\ref{tab:eagle-ablations} shows that synthetic training is more effective than using external data alone (1.37$\times$ vs. 1.19$\times$), underscoring the value of in-domain alignment. Combining both data sources further improves performance to 1.46$\times$, suggesting complementary gains in generalization across both tree and greedy decoding settings. While greedy decoding lowers MGL from 4.29 to 3.80, the reduced drafting cost and target-model verification overhead outweigh the decrease in acceptance rate, yielding a peak speedup of 1.78$\times$ under target-serving concurrency.

\paragraph{FP8 Calibration Performance.}
\label{sec:clip-rate} 
Static FP8 Post-Training Quantization is sensitive to the calibration data being used, so we measure the calibration performance on the mixed set along with its individual public and in-domain subsets. The E2E pass rate in Figure~\ref{fig:clip_rate} shows that we preserve performance with the mixed calibration set, and we analyze this using \textit{activation clip rate}: the fraction of test-time activations falling outside the per-tensor min/max bounds set during calibration. High clip rates indicate insufficient dynamic range and can cause silent regressions on long contexts. 

\paragraph{Clip Rate Regimes.} Figure~\ref{fig:clip_rate} shows two regimes. In the first regime, below 128 tokens, the public set has the lowest clip rate, since short internet snippets already cover a broad activation range. In the second regime, from 256 tokens onward, the mixed set dominates: at 2,048 tokens, it is $6.1\times$ lower than the public set and $1.4\times$ lower than the in-domain set. Production prompts in our system routinely exceed 1,000 tokens once tools, memory, and few-shot exemplars are injected, so the long-context regime is the operative one and motivates the mixed set as the deployed default.

\section{Related Work}
\label{sec:related}

Multi-agent architectures decompose problems across specialized agents~\cite{guo2024large,hong2024metagpt,wu2024autogen}, with coordination patterns including sequential pipelines and hierarchical orchestration~\cite{du2024improving}. \citet{chen2024more} observe that increasing agent calls yields diminishing returns without optimization. Our work shows how inference-level optimizations reduce per-call cost and increase achievable throughput in agentic workflows. Speculative decoding methods like EAGLE~\cite{li2024eagle,li2024eagle2,li2026eagle} accelerate generation via draft models, but their use in application-specific settings remains under-explored. We show the impact of mixed training data on draft acceptance rates and how it improves throughput. Post-training quantization compresses LLMs without retraining~\cite{frantar2023optq,xiao2023smoothquant,shen2024efficient}. FP8 quantization preserves quality better than integer formats for certain workloads~\cite{shen2024efficient,fishman2025scaling}. However, direct comparisons of public versus application-specific calibration data for FP8 PTQ are limited. Our results show that data mixture composition strongly affects quality preservation under compression.

\section{Practical Takeaways}
\label{sec:lessons}

The process of distilling complex agentic workflows into a compact, production-ready model yields several important findings. First, student model performance is largely bounded by the fidelity of the synthetic trajectories produced by the Agent Simulator, underscoring the critical role of high-quality synthetic data. Second, LoRA-based adaptation is necessary to preserve zero-shot generalization under evolving system prompts, whereas full-parameter fine-tuning degrades this capability. Finally, a layered optimization strategy that combines the CPT–SFT–DPO distillation pipeline with custom EAGLE drafters and FP8 quantization delivers a sustained 4.48× end-to-end speedup without measurable degradation in task intelligence. Additional details on these training methodologies and inference trade-offs are provided in Appendix~\ref{sec:appendix_takeaways}.

\section{Conclusion}
\label{sec:conclusion}

We describe an integrated optimization framework for our deployed, production-ready Multi-Agent System that achieves a 4.48$\times$ improvement in throughput with no measurable loss in quality. Our experiments show that FP8 post-training quantization requires mixed calibration to preserve performance, application-specific EAGLE drafters substantially outperform generic variants in token acceptance rates, and jointly optimized system components deliver multiplicative efficiency gains. These findings highlight the importance of holistic optimization and demonstrate that deployment-scale improvements in multi-agent LLM systems depend on coordinated data engineering, model adaptation, and system-level optimization.

\subsection*{Limitations}
The reported results are specific to our production multi-agent system in the automotive retail domain. The core principles of data alignment and multi-layer optimization generalize to other use cases, but exact performance gains vary by application. During training, our distillation pipeline relies heavily on a single model acting as both teachers and automated judges to simulate and verify synthetic traces. Any inherent biases, domain blind spots, or reasoning gaps in these models inevitably propagate to the student. This dependency requires manual intervention, including our hand-crafted DPO pairs, to correct business-logic edge cases that the automated judge misses. Additionally, generating and verifying hundreds of thousands of multi-turn conversational traces requires significant upfront computational resources. Also, the EAGLE drafter must be retrained whenever system prompts or business logic are updated, as these modifications induce distribution shifts that degrade draft acceptance rates. Finally, some of our inference optimizations are hardware-dependent. FP8 quantization requires native hardware support like NVIDIA Hopper GPUs, and falling back to higher precision execution severely reduces the reported throughput benefits.

\bibliography{custom}

\appendix

\section{Loss Formulations for Agentic Training Stages}
\label{sec:appendix_math_formulations}

In this section, we explicitly detail the mathematical loss functions minimized during the three sequential stages of our agentic model customization pipeline.

\subsection{Context-aware Continual Pretraining (CA-CPT)}
The loss objective for Stage 1 adjusts standard language modeling objectives by prefixing document $x \in \mathcal{X}_{\tt{CPT}}$ with its generated context token vector $C_x$:
\begin{equation}
\begin{aligned}
\mathcal{L}_{\tt{CA}-\tt{CPT}}(\theta) = -\underset{x \sim \mathcal{X}_{\tt{CPT}}}{\mathbb{E}} \Bigl[ \sum_{t} \log P_{\theta}(x_t \mid x_{<t}; C_x) \Bigr].
\end{aligned}
\end{equation}

\subsection{Supervised Fine-Tuning (SFT)}
The loss minimized during Stage 2 optimizes the model parameters over the distribution of refined instruction-following synthetic traces $\mathcal{Y}_{\text{SFT}}$:
\begin{equation}
\begin{aligned}
\mathcal{L}_{\tt{SFT}}(\theta) = -\underset{(x,y) \sim \mathcal{Y}_{\text{SFT}}}{\mathbb{E}} \Bigl[ \log P(y \mid x; \theta) \Bigr].
\end{aligned}
\end{equation}

\subsection{Direct Preference Optimization (DPO)}
Stage 3 aligns model generations using the choice triples $(x, y^*, y) \sim \mathcal{Z}_{\tt{DPO}}$ via the native implicit reward optimization objective:
\begin{equation}
\begin{aligned}
\mathcal{L}_{\tt{DPO}}(\theta) = & -\underset{(x, y^*, y) \sim \mathcal{Z}_{\tt{DPO}}}{\mathbb{E}} \\
& \left[ \log \sigma \left( \beta \log \frac{\pi_\theta(y^* \mid x)}{\pi_{\theta}^{\tt{SFT}}(y^* \mid x)} \right. \right. \\
& \left. \left. - \beta \log \frac{\pi_\theta(y \mid x)}{\pi_{\theta}^{\tt{SFT}}(y \mid x)} \right) \right].
\end{aligned}
\end{equation}

\section{Training Details}
\label{sec:appendix_training_details}

% \subsection{Model Architecture and Initialization}
% We select the \textsc{Llama 3.1 8B Instruct} model (\texttt{BF16}) as the base for $\pi_\theta$. To ensure sufficient parameter capacity for automotive domain logic, we perform capacity augmentation using the Llama Pro block expansion technique \cite{wu2024llama}. We integrate 8 additional transformer layers, resulting in a student scale of approximately \texttt{10.1B} parameters, maintained entirely in \texttt{BF16}.

% \subsection{Sequential Training Pipeline (CPT, SFT, DPO)}
% The development of $\pi_\theta$ proceeds in three sequential distillation stages to transfer agentic capabilities from the teacher $\pi_T$. All stages utilize the \texttt{BF16} data format for numerical stability. The specific training hyperparameters for the alignment stages are detailed in Table~\ref{tab:hyperparams}.

% \paragraph{Stage 1: Continual Pre-training (CPT).} 
% We initialize $\pi_\theta^{(0)}$ and perform CPT to instill foundational domain knowledge using extensive public text corpora. Crucially, the synthetic conversational data generated by our agentic simulator is strictly excluded from the CPT phase, reserving it exclusively for downstream task-specific alignment. To prevent catastrophic forgetting of the base model's conversational skills, we apply sparse training via SNR analysis \cite{hartford2024spectrum} on the original \texttt{BF16} layers. We update only high-SNR parameters while performing full-parameter updates on the 8 new layers, resulting in the intermediate checkpoint $\pi_\theta^{\text{CPT}}$.

\subsection{Why Context Reduces Forgetting in CPT}

The key intuition behind Context-Aware CPT is that the first few tokens of a training sequence often produce disproportionately high loss because the model has little or no preceding context. In continual pretraining, this high initial loss can introduce noisy, high-variance gradients that are less reflective of the model's true domain knowledge gap and more reflective of uncertainty caused by insufficient context. Since the negative log-likelihood gradient with respect to the logits is $\nabla_{z_t}\mathcal{L}_t = p_t - y_t$, uncertain predictions at early positions can induce large and unstable updates, pulling shared parameters in inconsistent directions across samples. Such variance is especially harmful in continual learning, where stochastic updates can move the model away from parameter regions that preserve previously learned capabilities, thereby contributing to catastrophic forgetting \citep{winata2023overcoming}. For a document $x = (x_1,\ldots,x_{|x|})$, the standard CPT objective is
\begin{equation}
\mathcal{L}_{\tt{CPT}}(x;\theta)
=
-\sum_{t=1}^{|x|}
\log p_{\theta}(x_t \mid x_{<t}).
\end{equation}
The corresponding gradient can be decomposed by token position:
\begin{equation}
\begin{aligned}
\nabla_{\theta}\mathcal{L}_{\tt{CPT}}(x;\theta)
&=
\underbrace{
-\sum_{t=1}^{k}
\nabla_{\theta}\log p_{\theta}(x_t \mid x_{<t})
}_{\nabla_{\theta}\mathcal{L}_{\tt{early}}}
+ \\
& \underbrace{
-\sum_{t=k+1}^{|x|}
\nabla_{\theta}\log p_{\theta}(x_t \mid x_{<t})
}_{\nabla_{\theta}\mathcal{L}_{\tt{later}}}.
\end{aligned}
\end{equation}
Here, $\nabla_{\theta}\mathcal{L}_{\tt{early}}$ denotes the gradient contribution from the first $k$ tokens, while $\nabla_{\theta}\mathcal{L}_{\tt{later}}$ denotes the contribution from the remaining tokens. Because early tokens are predicted with limited context, $\nabla_{\theta}\mathcal{L}_{\tt{early}}$ can have higher variance and may dominate the update direction despite carrying weaker domain-specific signal.

\begin{figure}[!th]
        \centering
        \includegraphics[width=0.5\textwidth]{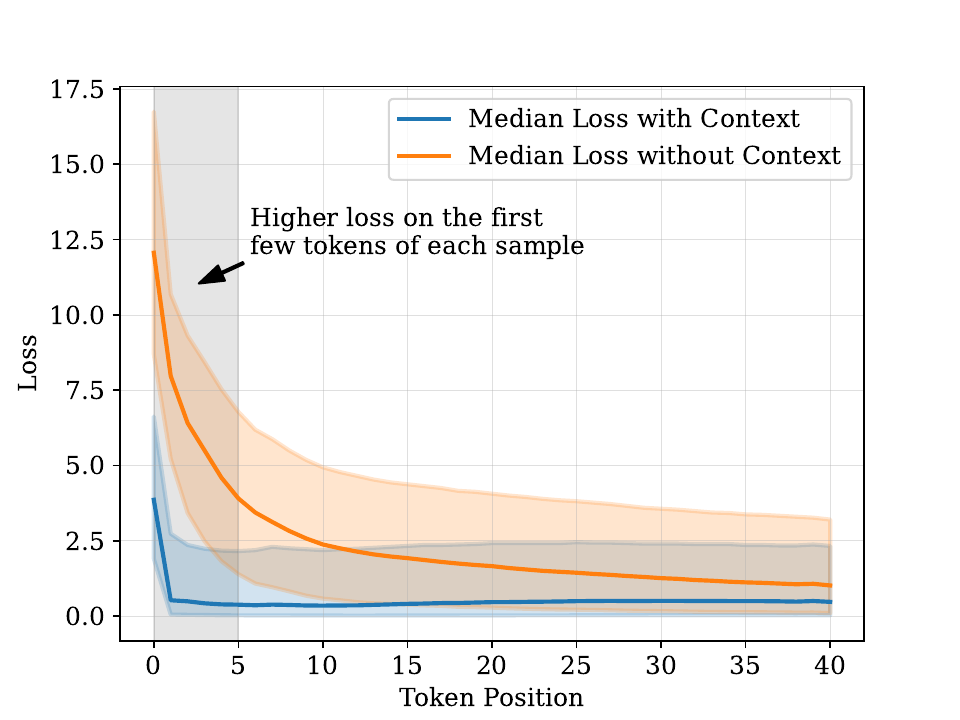}
        \caption{Loss and Token Position Across Domain Adaptation Datasets.}
        \label{fig:loss_token_pos}
\end{figure}

Context-Aware CPT reduces this instability by prepending each document with a sample-specific context $C_x$ and excluding the context tokens from the training loss. The resulting objective is
\begin{equation}
\mathcal{L}_{\tt{CA\text{-}CPT}}(x;\theta)
=
-\sum_{t=1}^{|x|}
\log p_{\theta}(x_t \mid  x_{<t}, C_x).
\end{equation}

By conditioning document tokens on $C_x$, the model receives a more informative prefix before predicting the original document content, reducing early-token uncertainty and improving the signal-to-noise ratio of CPT gradients. As a result, adaptation is driven by a cleaner and more contextually grounded training signal, allowing the model to absorb new domain knowledge while reducing destructive parameter drift and better balancing plasticity with stability.

% \paragraph{Stage 2: Supervised Fine-Tuning (SFT).} 
% Starting from $\pi_\theta^{\text{CPT}}$, we perform high-rank SFT using a mixture of public instructions and the internal synthetic agentic traces generated by our simulation pipeline. We apply LoRA \cite{hu2021lora} targeting all linear modules with a rank of $r=128$ and an alpha of 256. We employ neat packing to concatenate sequences within the 8,192 token cutoff. This stage yields the checkpoint $\pi_\theta^{\text{SFT}}$.

% \paragraph{Stage 3: Preference Alignment (DPO).} 
% The final stage applies Direct Preference Optimization (DPO) to $\pi_\theta^{\text{SFT}}$ to align the model to business requirements. Using the failure cases and judge-refined traces from our simulation pipeline, we build a targeted preference corpus consisting of triples $(x, y^*, y)$, where $y^*$ is a judge-refined "chosen" response and $y$ is a "rejected" response from the \texttt{70B} teacher $\pi_T$. We use a reduced LoRA rank of $r=64$ and a DPO penalty of $\beta=0.1$ to produce the final deployed model, $\pi_\theta^{\texttt{BF16}}$.

\subsection{Training Hyperparameters}

The detailed hyperparameters of CPT, Agentic Fine Tuning, Preference Tuning, and EAGLE can be found in Table \ref{tab:hyperparams}

\begin{table}[!th]
\centering
\resizebox{\columnwidth}{!}{%
\small
\begin{tabular}{lcccc}
\toprule
\textbf{Parameter} & \textbf{CPT ($\pi_\theta^{\text{CPT}}$)} & \textbf{SFT ($\pi_\theta^{\text{SFT}}$)} & \textbf{DPO ($\pi_\theta^{\text{final}}$)} & \textbf{EAGLE ($\pi_\theta^{\texttt{EAGLE}}$)} \\ 
\midrule
Precision & \texttt{BF16} & \texttt{BF16} & \texttt{BF16} & \texttt{BF16}\\
LoRA Rank ($r$) & -- & $128$ & $64$ & -- \\
LoRA Target & -- & All modules & All modules & --\\
Initial Learning Rate & $10^{-5}$ & $2.0 \times 10^{-5}$ & $5.0 \times 10^{-6}$ & $3.0 \times 10^{-4}$ \\
LR Scheduler & Cosine & Cosine & Cosine & Linear \\
Weight Decay & 0.1 &  -- & -- & --  \\
Warmup Ratio & $0.01$ & $0.05$ & $0.10$ & $0.01$\\
Epochs & 1 & $1$ & $3$ & $100$\\
Batch Size (per DP) & Varies (SEQ $\cdot$ GBS = 4M) & $1$ & $1$ & $8$\\
Gradient Accumulation & Varies (SEQ $\cdot$ GBS = 4M) & $1$ & $2$ & $1$\\
Max Context Length & $4k, 8k, 16k$ & $8k$ & $8k$ & $8k$\\
DPO Penalty ($\beta$) & -- & -- & $0.1$ & --\\ 
\bottomrule
\end{tabular}%
}
\caption{Hyperparameter configurations for the continual pretraining, supervised fine-tuning, preference alignment, and EAGLE stages of the student model $\pi_\theta$.}
\label{tab:hyperparams}
\end{table}

\subsection{Detailed Dataset Synthesis}
To curate a high-fidelity training corpus for the student model $\pi_{\theta}^{\texttt{BF16}}$, we utilize the LLM-driven agentic simulator pipeline detailed in Section~\ref{sec:distillation}. We simulate 7,172 conversations against the teacher system, producing an extensive corpus of 495,772 individual training traces spanning various intents and tool-calling behaviors. These traces constitute the Supervised Fine-Tuning ($\mathcal{Y}_{\text{SFT}}$) corpus. 

For the Preference Alignment (DPO) stage, we construct a specialized dataset ($\mathcal{Z}_{\text{DPO}}$) comprising 10,000 preference pairs. Each preference pair consists of a "chosen" (correct) response and a "rejected" (incorrect) response. The automated pipeline generates 9,000 of these pairs, where the LLM-as-a-Judge successfully flags mistakes in the teacher model's outputs and provides refined corrections. However, because the teacher model ($\pi_T$) is already highly optimized, relying solely on the automated judge to identify subtle logical errors proves challenging. To address this, we manually craft the remaining 1,000 hard-negative pairs. These manual pairs explicitly target known failure modes, complex business-logic boundary conditions, and specific scenarios where the model traditionally struggles and the automated judge fails to flag the error. 

To rigorously test the distilled model against unseen scenarios, we separately generate an internal evaluation set by running an additional 1,424 simulated conversations, yielding 8,848 distinct evaluation instances. Furthermore, to train the EAGLE drafter, which requires responses from the trained student model, we utilize the same agentic simulator pipeline with $\pi_{\theta}^{\texttt{BF16}}$ to generate 50,033 data points. Finally, for FP8 calibration, we sample 5,800 traces from this EAGLE dataset.

\subsection{Training Infrastructure}
Context-aware Continual Pretraining is conducted on $256$ nodes, each equipped with $8$ A100 GPUs, for a total of $2{,}048$ A100 GPUs. We train on approximately $5$T tokens using tensor parallelism with $\mathrm{TP}=8$, pipeline parallelism with $\mathrm{PP}=1$, and data parallelism with $\mathrm{DP}=256$. The micro-batch size is set to $1$, and we maintain a global batch size of approximately $4$M tokens by adjusting the gradient accumulation steps according to the sequence length, which ranges from $4$K to $16$K tokens.

Agentic fine-tuning (AFT) and DPO are performed on $16$ nodes with $8$ A100 GPUs per node, using Fully Sharded Data Parallelism (FSDP). High-bandwidth cross-node communication is enabled through Elastic Fabric Adapter (EFA) with RDMA support.  For FSDP training, we use the \texttt{FULL\_SHARD} strategy with CPU parameter offloading, backward prefetching, and transformer-based auto-wrapping.

For EAGLE drafter training, we utilize one AWS \texttt{p4d.24xlarge} instance and a standard data parallel setting. Until this stage, all model weights, activations, and gradients are maintained strictly in \texttt{BF16} precision to ensure numerical stability without sacrificing throughput.

\section{Multi-Agent Architecture and Deployment Details}
\label{sec:appendix_agent_architecture}

\subsection{Detailed Agent Roles}
\label{sec:appendix_agent_roles}

Our production Multi-Agent System ($M$) is a five-agent system for customer-facing reasoning tasks spanning intent understanding, plan generation, verification, and explanation. The agents collaborate as follows:

\begin{itemize}[leftmargin=*, parsep=0pt]
    \item \textbf{\textsc{Understander Agent}:} Interacts with the user to detect their core needs and gather all necessary information. It processes user utterances alongside chat history to maintain context, extracting essential state-level details and domain-specific entities.
    \item \textbf{\textsc{Planner Agent}:} Uses the extracted context to formulate an actionable strategy. It generates structured, executable action plans using provided tools ($\mathcal{T}$) while strictly complying with dealership business rules.
    \item \textbf{\textsc{Evaluator Agent}:} Operates as a critical safety guardrail by verifying the generated plans using both rule-based and LLM-based validation. If safety or logic violations are detected, it triggers a replanning loop, sending feedback to the \textsc{Planner Agent} for correction.
    \item \textbf{\textsc{Executor Agent}:} Once the \textsc{Evaluator Agent} validates the code, the \textsc{Executor Agent} securely runs the executable plan within an external environment and passes the results to the next agent.
    \item \textbf{\textsc{Explainer Agent}:} Finally, the \textsc{Explainer Agent} translates these executed plans and raw tool outputs into natural language explanations for the customer.
\end{itemize}

\subsection{Deployment Constraints and Performance Bottlenecks}
\label{sec:appendix_deployment_constraints}

All agents in our system invoke the same foundation model. The overarching goal for production deployment is to maximize throughput on AWS EC2 P5 (8$\times$ NVIDIA H100 80GB GPUs) while strictly adhering to sub-second end-to-end latency SLAs. 

Initial system profiling identified three major bottlenecks: 
\begin{enumerate}
    \item \emph{Cumulative latency} resulting from multiple sequential LLM calls per request, which compounds to multi-second delays.
    \item \emph{Memory footprint} constraints from serving large LLMs in BF16, which severely limits batch sizes and overall concurrent capacity.
    \item \emph{Generation cost}, which cannot be optimized through prefill optimization (such as prompt caching) alone. 
\end{enumerate}

While standard batching and quantization can partially address latency and memory constraints, the generation cost bottleneck requires a fundamentally different approach. Speculative decoding addresses this by verifying multiple speculated tokens per target-model forward pass, bypassing the traditional autoregressive bottleneck.

\section{Additional Serving Optimizations}
\label{sec:systems}

Beyond quantization and speculative decoding, we apply several systems-level changes to reduce end-to-end latency. Our serving baseline already includes many widely adopted systems-level optimizations, making it exceptionally strong and difficult to improve upon. The compounding gains from our proposed methods are achieved on top of this highly optimized foundation, which specifically includes:

\paragraph{Conditional Agent Invocation.} In production traffic, most Planner outputs are simple enough for deterministic evaluation. We measure plan complexity via Halstead complexity metrics and only invoke the LLM-based Evaluator when complexity exceeds a threshold; simple plans bypass the Evaluator entirely. This reduces total LLM calls per request.

% \paragraph{Structured Output Consolidation.} In the original implementation, our Multi-Agent System ($M$) issued separate LLM calls for each micro-decision (e.g., intent extraction, missing information identification, question formulation). We consolidate these into a single structured-output call using constrained decoding via xgrammar \cite{dong2025xgrammar}, producing a JSON object containing all required fields. This reduces per-request LLM invocations, improves prompt cache hit rates, and shifts more computation to the generation phase where EAGLE drafter provides speedup.

\paragraph{Continuous Batching.} Instead of waiting for the longest sequence in a static batch to finish, requests are scheduled at the iteration level. Completed requests are immediately replaced with new ones from the queue, maximizing GPU utilization.

\paragraph{Tensor Parallelism.} Model weights are sharded across multiple GPUs to distribute the memory footprint and compute load, significantly reducing time-to-first-token (TTFT) and per-token generation latency.

\paragraph{KV-Cache CPU Offloading.} To prevent Out-Of-Memory (OOM) errors and increase concurrent capacity, inactive KV-caches are dynamically swapped to host CPU memory and asynchronously prefetched back to VRAM when needed.

\begin{figure}[t]
    \centering
    \resizebox{0.5\textwidth}{!}{%
        % % =========================
% FILE: figures/optimization_stack.tikz
% =========================
\begin{tikzpicture}[
    font=\sffamily,
    layer/.style={
        rectangle,
        rounded corners=5pt,
        draw=black,
        thick,
        minimum width=11.8cm, % <-- WIDENED from 11cm to 11.8cm
        align=center,
        inner sep=4pt
    },
    badge/.style={
        rectangle,
        rounded corners=4pt,
        draw=black,
        thick,
        minimum width=1.4cm,
        minimum height=0.6cm,
        inner sep=2pt,
        font=\small\sffamily\bfseries,
        text=white,
        align=center
    },
    innerbox/.style 2 args={
        rectangle,
        rounded corners=3pt,
        draw=#2,
        fill=#1,
        thick,
        minimum width=2.8cm,
        text width=2.6cm, 
        minimum height=1.0cm,
        align=center,
        font=\small\sffamily,
        text=black,
        inner sep=3pt
    },
    pipearrow/.style={
        ->,
        line width=1.1pt,
        draw=black!55,
        >={Stealth[length=2.5mm,width=2mm]}
    },
    caption/.style={
        font=\footnotesize\sffamily\itshape,
        text width=11.2cm, % <-- WIDENED to match the new layer width
        align=center,
        inner sep=0pt
    },
    compound/.style={
        circle,
        draw=black!55,
        thick,
        fill=white,
        minimum size=0.55cm,
        inner sep=0pt,
        font=\small\sffamily\bfseries,
        text=black
    }
]

% =====================================================
% LAYER 4 — End-to-End Impact
% =====================================================
\node[layer, fill=orange!18, minimum height=2.4cm] (layer4) {};  
% Changed xshift to 0.6cm to pull the text slightly left
\node[badge, fill=orange!70!black] (b4) at ([xshift=0.6cm, yshift=-0.55cm]layer4.north west) {Layer 4};
\node[anchor=west, font=\normalsize\sffamily\bfseries] at ([xshift=0.20cm]b4.east)
    {End-to-End Impact};
\node[innerbox={orange!8}{orange!55!black}] at ([xshift=-3.6cm, yshift=-0.21cm]layer4.center)  
    {{\color{red}\ding{116}}\,\textbf{P90 Latency}\\[-1pt]\scriptsize per-query tail};
\node[innerbox={orange!8}{orange!55!black}] at ([xshift=0cm,    yshift=-0.21cm]layer4.center)
    {{\color{green!50!black}\ding{115}}\,\textbf{Throughput}\\[-1pt]\scriptsize QPS / GPU};
\node[innerbox={orange!8}{orange!55!black}] at ([xshift=3.6cm,  yshift=-0.21cm]layer4.center)
    {{\color{red}\ding{116}}\,\textbf{Cost / Query}\\[-1pt]\scriptsize \$ per request};
\node[caption, anchor=south, yshift=0.10cm] at (layer4.south)
    {Speedups compound multiplicatively across layers.};

% =====================================================
% LAYER 3 — Throughput Density (FP8)
% =====================================================
\node[layer, fill=blue!16, minimum height=2.4cm, below=0.3cm of layer4] (layer3) {};  
\node[badge, fill=blue!60!black] (b3) at ([xshift=0.6cm, yshift=-0.55cm]layer3.north west) {Layer 3};
\node[anchor=west, font=\normalsize\sffamily\bfseries] at ([xshift=0.20cm]b3.east)
    {Throughput Density \,\footnotesize\textemdash{} FP8 (W8A8) PTQ};
\node[innerbox={blue!8}{blue!55!black}] at ([xshift=-3.6cm, yshift=-0.21cm]layer3.center)  
    {{\color{red}\ding{116}}\,\textbf{Memory}\\[-1pt]\scriptsize $\sim$2$\times$ smaller};
\node[innerbox={blue!8}{blue!55!black}] at ([xshift=0cm,    yshift=-0.21cm]layer3.center)
    {{\color{green!50!black}\ding{115}}\,\textbf{Compute}\\[-1pt]\scriptsize $\sim$2$\times$ TC throughput};
\node[innerbox={blue!8}{blue!55!black}] at ([xshift=3.6cm,  yshift=-0.21cm]layer3.center)
    {{\color{green!50!black}\ding{52}}\,\textbf{Quality}\\[-1pt]\scriptsize per-tensor scaling};
\node[caption, anchor=south, yshift=0.10cm] at (layer3.south)
    {Half the memory, double the tensor-core throughput, no quality loss.};

% =====================================================
% LAYER 2 — Per-Call Generation Latency (EAGLE)
% =====================================================
\node[layer, fill=green!16, minimum height=2.6cm, below=0.3cm of layer3] (layer2) {};  
\node[badge, fill=green!45!black] (b2) at ([xshift=0.6cm, yshift=-0.55cm]layer2.north west) {Layer 2};
\node[anchor=west, font=\normalsize\sffamily\bfseries] at ([xshift=0.20cm]b2.east)
    {Per-Call Generation Latency \,\footnotesize\textemdash{} EAGLE Spec.\ Decoding};

\node[innerbox={green!8}{green!50!black}] (d) at ([xshift=-3.4cm, yshift=-0.15cm]layer2.center)  
    {\textbf{Draft model}\\[-1pt]\scriptsize proposes $k$ tokens};
\node[innerbox={green!8}{green!50!black}] (v) at ([xshift=0cm,    yshift=-0.15cm]layer2.center)
    {\textbf{Verifier target LLM}\\[-1pt]\scriptsize parallel check};
\node[innerbox={green!8}{green!50!black}] (a) at ([xshift=3.4cm,  yshift=-0.15cm]layer2.center)
    {\textbf{Accepted tokens}\\[-1pt]\scriptsize $\bar{k}>1$ per step};
\draw[pipearrow] (d) -- (v);
\draw[pipearrow] (v) -- (a);

\node[caption, anchor=south, yshift=0.10cm] at (layer2.south)
    {Amortizes target-model decode cost; fewer sequential steps per token.};

% =====================================================
% LAYER 1 — Systems-Level Call Reduction
% =====================================================
\node[layer, fill=purple!16, minimum height=2.6cm, below=0.3cm of layer2] (layer1) {};  
\node[badge, fill=purple!60!black] (b1) at ([xshift=0.6cm, yshift=-0.55cm]layer1.north west) {Layer 1};
\node[anchor=west, font=\normalsize\sffamily\bfseries] at ([xshift=0.20cm]b1.east)
    {Systems-Level Call Reduction};

\node[innerbox={purple!8}{purple!55!black}, text width=2.6cm, minimum width=3.2cm, minimum height=1.1cm] at ([xshift=-3.75cm, yshift=-0.20cm]layer1.center)                                                                          
      {\textbf{Conditional Agent Invocation}};                                               
  % \node[innerbox={purple!8}{purple!55!black}, text width=2.6cm, minimum width=3.2cm, minimum height=1.1cm] at ([xshift=-1.25cm, yshift=-0.20cm]layer1.center)                                                                        
  %     {\textbf{Structured Output Consolidation}};
  \node[innerbox={purple!8}{purple!55!black}, text width=2.6cm, minimum width=3.2cm, minimum height=1.1cm] at ([xshift=0cm,     yshift=-0.20cm]layer1.center)                                                                          
      {\textbf{Prompt-Cache Reuse}};                                                         
  \node[innerbox={purple!8}{purple!55!black}, text width=2.6cm, minimum width=3.2cm, minimum height=1.1cm] at ([xshift=3.75cm,  yshift=-0.20cm]layer1.center)
      {\textbf{Continuous Batching}};

\node[caption, anchor=south, yshift=0.10cm] at (layer1.south)
    {Removes redundant calls; raises cache hits; lifts GPU utilization.};

% =====================================================
% LAYER 0 — Base Inference
% =====================================================
\node[layer, fill=gray!22, minimum height=2.1cm, below=0.3cm of layer1] (layer0) {};  
\node[badge, fill=black!55] (b0) at ([xshift=0.6cm, yshift=-0.55cm]layer0.north west) {Layer 0};
\node[anchor=west, font=\normalsize\sffamily\bfseries] at ([xshift=0.20cm]b0.east)
    {Base Inference \,\footnotesize\textemdash{} unoptimized baseline};

\node[innerbox={white}{black!55}] at ([xshift=-3.6cm, yshift=-0.4cm]layer0.center)  
    {\textbf{BF16} precision};
\node[innerbox={white}{black!55}] at ([xshift=0cm,    yshift=-0.4cm]layer0.center)
    {$\pi_{\theta}^{\text{BF16}}$ model};
\node[innerbox={white}{black!55}] at ([xshift=3.6cm,  yshift=-0.4cm]layer0.center)
    {\textbf{1} call / agent step};

\end{tikzpicture}
    }
    \caption{\textbf{Optimization Stack.} The four optimization layers (L1–L4) yield compounding performance gains over the unoptimized baseline (L0).}
    \label{fig:optimization_stack}
\end{figure}
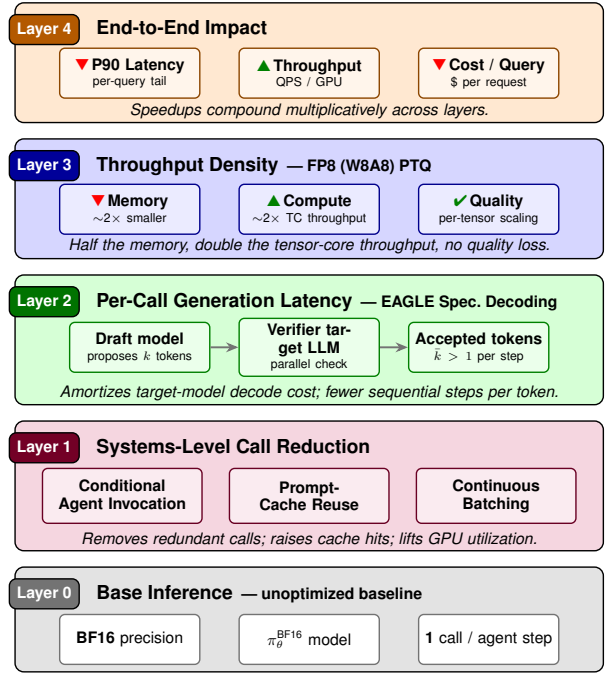

Figure~\ref{fig:optimization_stack} illustrates the different layers of our complete optimization stack.

\section{Detailed Discussions and Practical Takeaways}
\label{sec:appendix_takeaways}

The process of distilling complex agentic workflows into a smaller, production-ready model yields several critical insights regarding data generation, training methodologies, and inference optimization.

\paragraph{The Crucial Role of the Agent Simulator.} 
We find that the quality of the student model is entirely bottlenecked by the fidelity of the synthetic data. Building an effective Agent Simulator requires rigorous, manual optimization of its governing prompts to ensure it accurately mirrors the distribution and nuances of real-world production conversations. Investing time in a high-quality simulator is paramount; without it, the downstream distillation process will simply reinforce unrealistic interaction patterns.

\paragraph{Preserving Prompt Adherence with LoRA.} 
In a live production environment, business requirements frequently evolve. Product teams regularly need to introduce new tool APIs, alter business logic, or modify the user experience. Consequently, the distilled student model must remain highly adaptable to system prompt updates. We initially experiment with full-parameter fine-tuning. Although it achieves comparable baseline performance, the fully fine-tuned model severely overfits the specific prompt structures seen during training, losing its ability to generalize or adapt to new instructions. Conversely, applying LoRA successfully preserves the foundation model's innate zero-shot adaptability. LoRA allows the model to learn the required domain expertise while remaining responsive to subsequent prompt modifications, which is a mandatory requirement for maintaining a dynamic production system.

\paragraph{The Necessity of Preference Alignment.} 
Supervised Fine-Tuning (SFT) alone is insufficient for achieving production-grade reliability. While SFT successfully instills the general tool-calling formats and conversational tone, Direct Preference Optimization (DPO) is essential for addressing complex boundary conditions. By explicitly contrasting successful outputs against failure modes, DPO effectively corrects nuanced logical errors and edge cases where even the high-parameter teacher model occasionally struggles.

\paragraph{Stacking Inference Optimizations.} 
In a large design space of optimizations~\cite{zhen2025taming, zhao2025optimizing}, we discover the importance of stacking optimization combinations that provide durable acceleration while preserving intelligence. At the systems level, call reduction techniques such as conditional agent invocation, prompt-cache reuse, and continuous batching remove redundant calls and raise GPU utilization and are included in evaluating all baselines. CPT–SFT–DPO Distillation from $\pi_T$ to $\pi_{\theta}^{\texttt{BF16}}$ provides a 2.33$\times$ E2E speedup. Trained EAGLE drafters and W8A8-FP8 further durably accelerate $\pi_{\theta}^{\texttt{BF16}}$ by 1.92$\times$, stacking to produce $\pi_{\theta}^{\texttt{EAGLE+FP8}}$ with a 4.48$\times$ speedup while staying under latency SLOs (Service Level Objectives). We also find that, across optimization phases, proper public and in-domain data mixtures are critical to prevent catastrophic forgetting during distillation, ensure robust FP8-W8A8 calibration, and enable strong acceptance lengths in EAGLE drafter training. Finally, careful tradeoffs such as choosing greedy speculation as opposed to tree speculation improve latency and throughput even when resulting in a lower speculative MGL.

\paragraph{Draft Model Quantization and Greedy Decoding.}
\label{sec:appendix_takeaways_model_quantization}
Quantizing the 250M EAGLE drafter with a mixed calibration set under greedy decoding moves speedup from 4.16$\times$ ($\pi_{\theta}^{\texttt{EAGLE}}$ (C)) to 4.48$\times$ ($\pi_{\theta}^{\texttt{EAGLE+FP8}}$), reflecting higher draft throughput at an unchanged MGL (3.80 tokens) as latency drops from 0.96s to 0.92s and QPS rises from 6.07 to 6.54.

Standard EAGLE uses tree-structured draft expansion, generating multiple candidate continuations per step that are verified in a single forward pass of the target model via tree attention. Greedy draft decoding (argmax sampling) generates a single candidate chain per step, reducing per-step compute at the cost of lower acceptance length. On $\pi_{\theta}^{\texttt{EAGLE}}$ (Combined data), switching from tree to greedy decoding increases the speedup from 3.40$\times$ to 4.16$\times$ (latency 1.19s $\rightarrow$ 0.96s, QPS 4.96 $\rightarrow$ 6.07) despite MGL dropping from 4.29 to 3.80 tokens. The same trend holds for the Synthetic adapter: $\pi_{\theta}^{\texttt{EAGLE}}$ (Synthetic data) tree decoding yields 3.19$\times$, while greedy yields 3.77$\times$.

At low concurrency, greedy slightly increases latency because acceptance breadth dominates the cost of each verification pass. At high concurrency, which is the operating regime in Table~\ref{tab:quantization-eagle}, draft throughput becomes the bottleneck and greedy yields latency reduction. Motivated by the strong performance of \texttt{EAGLE+FP8} in the greedy regime in our ablations, we make upstream contributions to vLLM to natively enable \texttt{EAGLE+FP8}, making it available to the community.

% \skcomment{We also want to add tree ablations here: (depth, topk, total tokens) and compare it with greedy decoding. Add ablation results around input concurrency, tree/chain drafting.}

% \skcomment{I think we can also expt with and results of training own LM Head and embedding layer for eagle drafter.}

% \kbcomment{We should mention our OSS contribution to VLLM to enable FP8+EAGLE}

\end{document}